\documentclass{article} %
\usepackage{iclr2025_conference,times}

\usepackage{hyperref}
\usepackage{url}

\usepackage{lineno}
\usepackage{tikz}
\usetikzlibrary{positioning}
\usetikzlibrary{arrows}
\usepackage{caption}
\usepackage{subcaption}
\usepackage{titlesec}

\usepackage{graphicx} 
\usepackage{tikz}
\usepackage{float}
\usepackage{booktabs}

\usepackage{enumitem}
\usepackage{booktabs}
\usepackage{multirow}
\usepackage{subcaption}
\usepackage{hyperref}
\usepackage{url}
\usepackage{algorithm}
\usepackage{algpseudocode}
\usepackage{amsmath}
\usepackage{nicefrac}
\usepackage{amssymb}
\usepackage{pifont}
\newcommand{\cmark}{\ding{51}}%
\newcommand{\xmark}{\ding{55}}%

\usepackage{bbm}
\usepackage{amsthm}
\usepackage{xfrac}

\usepackage{mathtools}

\usepackage{cleveref}
\usepackage{wrapfig}
\usepackage{adjustbox}

\usepackage{amsmath,amsfonts,bm}

\def\eqref#1{equation~\ref{#1}}

\def\1{\bm{1}}

\DeclareMathAlphabet{\mathsfit}{\encodingdefault}{\sfdefault}{m}{sl}
\SetMathAlphabet{\mathsfit}{bold}{\encodingdefault}{\sfdefault}{bx}{n}

\newcommand{\E}{\mathbb{E}}

\usepackage{hyperref}
\usepackage{url}
\usepackage{graphicx}
\usepackage{makecell}
\usepackage{soul}
\usepackage{afterpage}

\hypersetup{
   breaklinks=true,   %
   colorlinks=true,   %
}
\definecolor{red}{HTML}{ca0020}
\definecolor{lightred}{HTML}{f4a582}
\definecolor{lightblue}{HTML}{92c5de}
\definecolor{green}{HTML}{008837}
\definecolor{blue}{HTML}{2c7bb6}

\def\ptarget{p_\text{target}}
\def\ptilde{\tilde{p}_\text{target}}

\title{No Trick, No Treat: \\Pursuits and Challenges Towards \\Simulation-free Training of Neural Samplers}

\author{Jiajun He$^{*, 1}$\thanks{Equal Contribution. Corresponding to \texttt{jh2383@cam.ac.uk}, \texttt{yuanqidu@cs.cornell.edu}},
Yuanqi Du$^{*, 2}$,
Francisco Vargas$^{1,3}$,
Dinghuai Zhang$^{4}$, \\
\AND
Shreyas Padhy$^{1}$,
RuiKang OuYang$^{1}$,
Carla Gomes$^{2}$,
José Miguel Hernández-Lobato$^{1}$
\Note $^1$University of Cambridge, $^2$Cornell University, $^3$Xaira Therapeutics, $^4$Microsoft Research
}

 \iclrfinalcopy
\begin{document}

\maketitle

\begin{abstract}
We consider the \textit{sampling} problem, where the aim is to draw samples from a distribution whose density is known only up to a normalization constant. 
Recent breakthroughs in generative modeling to approximate a high-dimensional data distribution have sparked significant interest in developing neural network-based methods for this challenging problem. 
However, neural samplers typically incur heavy computational overhead due to simulating trajectories during training. 
This motivates the pursuit of \textit{simulation-free} training procedures of neural samplers. 
In this work, we propose an elegant modification to previous methods, which allows simulation-free training with the help of a time-dependent normalizing flow. 
However, it ultimately suffers from severe mode collapse. 
On closer inspection, we find that nearly all successful neural samplers rely on Langevin preconditioning to avoid mode collapsing. 
We systematically analyze several popular methods with various objective functions and demonstrate that, in the absence of  Langevin preconditioning, most of them fail to adequately cover even a simple target. 
Finally, we draw attention to a strong baseline by combining the state-of-the-art MCMC method, Parallel Tempering (PT), with an additional generative model to shed light on future explorations of neural samplers.
\end{abstract}

\section{Introduction}
Sampling is a fundamental task in statistics, with broad applications in Bayesian inference, rare event sampling, and molecular simulation~\citep{box2011bayesian,tuckerman2023statistical,dellago2002transition,dudoob}.
Consider a target distribution with the following density function:
\begin{equation}
\ptarget(x) = \frac{\ptilde(x)}{Z}, \quad Z = \int_{\Omega} \ptilde(x) dx,
\end{equation}
where $\ptilde(x)$ is the unnormalized density which we can evaluate for a given $x$, and $Z$ is an unknown normalization factor.
We aim to generate samples following $\ptarget$.
These samples can be used to estimate the normalization factor or the expectation over some test functions.
\par
A ``standard" solution to this problem is Markov chain Monte Carlo (MCMC), which runs a Markov chain whose invariant density is $\ptarget$. 
Building on top of MCMC, various advanced sampling techniques have been developed, with the most efficient methods including Parallel Tempering (PT)~\citep{swendsen1986replica,earl2005parallel}, Annealed Importance Sampling (AIS)~\citep{neal2001annealed}, and Sequential Monte Carlo (SMC)~\citep{doucet2001introduction}.
However, MCMC-based approaches typically suffer from slow mixing time and dependency between samples. 
\par
A growing trend of research directions therefore focus on the learned neural sampler, e.g.,~\citep{noe2019boltzmann}, where we train a neural network to amortize the sampling procedure. 
Initial attempts studied normalizing flows (NFs) and used them as proposals for importance sampling (IS)~\citep{noe2019boltzmann,midgleyflow}. 
Later, diffusion and control-based samplers gained notable attention ~\citep{zhangpath,doucet2022score,vargasdenoising,berneroptimal,vargas2024transport,albergo2024nets} due to their success in generative modeling \citep{ho2020denoising,songscore,karras2022elucidating}.
These methods start with an easy-to-sample distribution (e.g., Gaussian) and evolve them through a stochastic differential equation (SDE) or ordinary differential equation (ODE).
However, despite significant progress, these approaches typically require \emph{simulating} the entire trajectory to evaluate the training objective. 
For instance, the most common objective - the reverse KL divergence between the model path measure and the target path measure - generally necessitates simulating the full trajectory for every sample and backpropagating through it. 
This leads to substantial memory consumption and slows down the training process.
\par
To this end, various objectives have been proposed to reduce computational costs. 
Some off-policy objectives enable detaching the gradient from the simulation~\citep{richterimproved}, while others involve simulating only a partial path~\citep{zhangdiffusion}.
The ultimate goal, however, is to design a sampler and training objective that can be optimized without any trajectory simulation following lessons learned from diffusion and flow matching models~\citep{ho2020denoising,songscore,lipmanflow}.
While appealing, we argue that most current approaches are not well-suited for such a design. 
This obstacle stems not only from how to modify the objective formulation for simulation-free evaluation but also from these approaches' reliance on tricks in network parameterization and sampling procedures that are not compatible with simulation-free training - most notably, the Langevin preconditioning, first proposed by \citet{zhangpath}. 
Through a simple example, we demonstrate that even with the same objective and a mode covering initialization, simulation-free training leads to significant mode collapse. 
We attribute this failure to the absence of the Langevin preconditioning in the simulation-free training pipeline.
To further support this claim, we provide ablation studies, showing that most current approaches struggle without the Langevin preconditioning. 
This observation highlights critical caveats and considerations that must be addressed in future work aimed at developing training-free objectives and pipelines.
\par
Running simulations with the Langevin preconditioning also poses a new challenge: 
simulation during training greatly increases the number of evaluations of the target density, which can be prohibitively expensive in some applications.
Consequently, it remains unclear whether these approaches are efficient compared to directly running MCMC and fitting a diffusion sampler post-hoc.
To investigate this, we compare the samplers with a state-of-the-art MCMC method, Parallel Tempering \citep[PT, ][]{PhysRevLett.57.2607,earl2005parallel}.
We find that PT serves as a remarkably strong baseline that should not be overlooked. 
\par
In summary, our main contributions are as follows:
(1) We provide a systematic review of current samplers, focusing on classifying different approaches by their underlying process and objectives. 
(2) We propose a simple direction for achieving it using Normalizing Flows. 
    Unfortunately, this attempt does not perform as desired, which we attribute to the absence of Langevin preconditioning widely applied in other neural samplers.
(3)  We investigate the influence of Langevin preconditioning. 
    Our findings reveal that most approaches fail when the sampler is not parameterized with the gradient of the target density. 
    This indicates critical caveats and considerations that should be addressed in developing simulation-free approaches.
(4) Finally, we compare several diffusion neural samplers with PT, and find that they lag significantly behind the results obtained from running traditional MCMC methods and fitting a diffusion model post hoc. 
    This highlights key challenges and critical considerations for enhancing the practicality of neural samplers in future work.

\par

\vspace{-5pt}
\section{Review of diffusion and controlled samplers}\label{sec:review}
\vspace{-5pt}
Before discussing the potential design of a simulation-free training approach, we first present a systematic review of current diffusion and controlled-based samplers in this section.
Despite the abundance of existing approaches, most samplers can be broadly categorized based on their sampling processes and training objectives:\vspace{-3pt}
\begin{enumerate}[leftmargin=*]
    \item \textbf{Sampling processes}. We can write the sampling process as follows:
    \begin{align}\label{eq:x}
        d X_t = \left [\mu_t(X_t) + \sigma_t^2 b_t(X_t) \right] dt + \sigma_t \sqrt{2} dW_t, \quad X_0 \sim p_\text{prior},
    \end{align}
    fixing or learning $\{\mu_t, \sigma_t, b_t, p_\text{prior}\}$ results in different sampling strategies. 
    Broadly, there are three main types of processes:
    \begin{itemize}[leftmargin=*]
        \item  \emph{time-reversal sampler}: the first involves training \Cref{eq:x} to approximate the time-reversal of a target process that begins with the target  $p_\text{target}$ and evolves toward a tractable distribution such as $p_\text{prior}$. 
        The target process is typically designed with a tractable drift term to ensure that its terminal density (approximately) converges to $p_\text{prior}$, with common choices including variance-preserving (VP) and variance-exploding (VE) SDEs and pinned Brownian motion (PBM).
        This category includes methods like PIS \citep{zhangpath,vargas2021bayesian}, DDS \citep{vargasdenoising}, DIS \citep{berneroptimal}, and iDEM \citep{akhounditerated}.
        \item \emph{escorted transport sampler}: the second trains \Cref{eq:x} to transport between a sequence of prescribed marginal densities $\pi_t$, typically defined by interpolation between $p_{\text{prior}}$ and $p_{\text{target}}$, with $\pi_0 = p_{\text{prior}}$ and $\pi_T = p_{\text{target}}$.
    Representative methods include Escorted Jarzynski \citep{vaikuntanathan2011escorted},  CMCD \citep{vargas2024transport}, NETS / PINN-based transport \citep{mate2023learning,albergo2024nets}, LFTS \citep{tian2024liouville}, etc.
    \item \emph{annealed variance reduction sampler}: Similar to the escorted transport sampler, these approaches prespecify an annealed target $\pi_t$  and set $b_t=0$ and  $\mu_t = \nabla \ln \pi_t$ just like the proposal in AIS \citep{neal2001annealed,jarzynski1997nonequilibrium}, the forward process remains fixed so no guidance/escorting is learned. 
    However, one approximates the reversal of this forward proposal so that the Radon-Nikodym derivative (RND) between the time-reversal and the forward proposal has a low variance, allowing a more efficient importance sampling.
    This category includes methods like AIS \citep{neal2001annealed,jarzynski1997nonequilibrium}, MCD \citep{doucet2022score,zhang2021some,hartmann2019jarzynski}, LDVI \citep{geffner2023langevin}, among others. 
    \end{itemize}
    \begin{table}[t]
\centering
\caption{Properties of different sampling processes. 
}\label{tab:process_property}\vspace{-5pt}
\begin{tabular}{llccc}
\hline
\multicolumn{2}{l}{\multirow{2}{*}{\textbf{Underlying process}}} & \multicolumn{3}{c}{\textbf{Properties}} \\
\multicolumn{2}{l}{} & \small{non-ergodicity} & \small{arbitrary $p_\text{prior}$} & \small{no mode switching} \\ \hline
\multicolumn{2}{l}{\textbf{Reversal of VP/VE SDE}} & \xmark & \xmark & \cmark \\
\multicolumn{2}{l}{\textbf{Reversal of PBM}} & \cmark & \xmark & \cmark \\
\multicolumn{2}{l}{\textbf{Escorted Transport}  (geom. interpolate)}  & \cmark & \cmark & \xmark \\\hline
\end{tabular}\vspace{-7pt}
\end{table}

  We compare the properties of different underlying processes in \Cref{tab:process_property}, including ergodicity (i.e., whether the sampler can mix within a finite number of steps \citep{albergo2023stochastic,huang2021schr,zhangpath,vargas2021bayesian,grenioux2024stochastic}), flexibility on the choice of prior, and the ``smoothness" \citep{chemseddine2024neural,woodard2009sufficient,tawn2020weight,syed2022non,phillips2024particle} of the induced flow
  (i.e. the mass teleportation problem, also known as mode switching).
    \item \textbf{Training objectives}.
    There are mainly two families of objectives:
    \begin{itemize}[leftmargin=*]
        \item \emph{path measure alignment}: the first one aligns the path measure induced by the sampling process, i.e., the SDE starting from $p_\text{prior}$, with another process starting from the target distribution $p_\text{target}$ and traversing in reverse. Common objectives include KL divergence \citep{zhangpath,vargas2021bayesian,vargasdenoising,doucet2022score,lahlou2023theorycontinuousgenerativeflow,berneroptimal,vargas2024transport}, log-variance divergence \citep{richterimproved}, the (sub-)trajectory balance objective~\citep{zhangdiffusion}, and detailed balance objective~\citep{bengio2021flow}. 
        \item \emph{marginal alignment}:  this approach aims to align the drift term or vector field of the sampling process with a prescribed target, ensuring that the marginal distributions of the generated samples closely follow the desired trajectory at each time step.
        Common objectives in this category include the physics-informed neural network (PINN) loss \citep{sun2024dynamical,albergo2024nets}, action matching loss \citep{albergo2024nets}, and score matching with importance sampling \citep{akhounditerated}.
    \end{itemize}
   We compare the properties of different objectives in \Cref{tab:obj_property}. 
   Specifically, we assess whether they support off-policy training, can be computed without simulation, require the costly calculation of network divergence, and ensure unbiasedness. 
   
   We note that the \emph{simulation-free training} can relate to several concepts in the neural sampler literature: (1) training without using MCMC, (2) detaching gradients on samples when evaluating trajectory-based objectives, and (3) evaluating objectives at any time step without simulating the trajectory.
  In this paper, we formally define simulation-free training as training with an objective that can be evaluated without simulating any ODE or SDE, aligning with the principles of diffusion and flow matching methods.
\end{enumerate}

\begin{table}[t]
\centering
\caption{Properties of different objectives. 
*KL divergence does not support simulation-free training in general.
However, it can be calculated without simulation for some special cases. 
We will provide an example later in \Cref{sec:sim_free}.
**KL divergence and log-variance divergence typically do not require computing the divergence. However, \citet{richterimproved} proposed objectives for neural samplers based on the general Schrödinger Bridge that requires computing this divergence.
}\label{tab:obj_property}
\begin{tabular}{@{}lcccc@{}}
\toprule
\multirow{2}{*}{\textbf{Objective}} & \multicolumn{4}{c}{\textbf{Properties}} \\
 & off-policy & sim-free & div-free & unbiased \\ \midrule
\textbf{KL} & \xmark & \xmark(\cmark*) & \cmark(\xmark**) & \cmark \\
\textbf{LV} & \cmark & \xmark & \cmark(\xmark**) & \cmark \\
\textbf{TB/STB} & \cmark & \xmark & \cmark & \cmark \\
\textbf{DB} & \cmark & \xmark & \cmark & \cmark \\
\textbf{PINN} & \cmark & \cmark & \xmark & \cmark \\
\textbf{AM} & \cmark & \xmark & \cmark & \cmark \\
\textbf{SM w. IS} & \cmark & \cmark & \cmark & \xmark \\
\bottomrule
\end{tabular}\vspace{-7pt}
\end{table}

Combining different underlying processes and objectives, we will recover many common neural samplers.
In the following, we briefly explain their design and categorize them in \Cref{tab:neural_samplers}. 
We include more details in \Cref{appendix:review}.
    \begin{enumerate}[label=({{\arabic*}}), leftmargin=*]
        \item Path Integral Sampler \citep[PIS,][]{zhangpath} and concurrently \citep[NSFS, ][]{vargas2021bayesian}: 
        PIS fixes $p_\text{prior} = \delta_0,  \sigma_t=1/\sqrt{2}$ and learns a network $f_\theta(\cdot) =\mu_t(\cdot) + \sigma_t^2 b_t(\cdot)$ so that \Cref{eq:x} approximate the time-reversal of the following SDE (Pinned Brownian Motion):
        \begin{align}\label{eq:pis_y}
             d Y_t = -\frac{Y_t}{T-t} dt + dW_t, \quad Y_0 \sim \ptarget.
        \end{align}
        We define \Cref{eq:pis_y} as the time-reversal of \Cref{eq:x} when $Y_t \sim X_{T-t}$.
        The network is learned by matching the reverse KL \citep{zhangpath,vargas2021bayesian} or log-variance divergence \citep{richterimproved} between the sampling and the target process.
        \par
        Diffusion generative flow samplers~\citep[DFGS,][]{zhangdiffusion} time-reversal the same pinned Brownian motion but with a new introduction of local objectives including detailed balance and (sub-)trajectory balance which has been shown equivalent to the log-variance objective with a learned baseline rather than a Monte Carlo (MC) estimator~\citep{nusken2021solving}.  
        \item Denoising Diffusion Sampler \citep[DDS,][]{vargasdenoising} and time-reversed Diffusion Sampler \citep[DIS,][]{berneroptimal}: 
        both DDS and DIS fix $\mu_t(X_t, t) = \beta_{T-t} X_t, \sigma_t = v \sqrt{\beta_{T-t}}, p_\text{prior} = \mathcal{N}(0, v^2I)$, and learn a network $f_\theta(\cdot, t) = b_t(\cdot, t)/2 $ so that \Cref{eq:x} approximates the time-reversal of the VP-SDE:
        \begin{align}\label{eq:dds_y}
            dY_t = -\beta_t Y_t dt + v \sqrt{2\beta_{t}} dW_t, \quad  Y_0 \sim \ptarget.
        \end{align}
        In an optimal solution, $f_\theta$ will approximate the score $f_\theta(\cdot, t) \approx \nabla\log p_{T-t}(\cdot)$, where $ p_{t}(X) = \int  \mathcal{N}(X|\sqrt{1-\lambda_t}Y, v^2\lambda_t I) \ptarget(Y)dY$ and $\lambda_t = 1-\exp(-2\int_0^t  \beta_s ds)$.
        Similar to PIS, the network can be trained either with reverse KL divergence or log-variance divergence.
        \item Iterated Denoising Energy Matching \citep[iDEM,][]{akhounditerated}: iDEM fixes $\mu_t(X_t, t) = 0, p_\text{prior} = \mathcal{N}(0, T^2 I)$, and learns a network $f_\theta(\cdot, t) = b_t(\cdot, t) / 2$ to approximate the score  $f_\theta(\cdot, t) \approx \nabla \log p_{T-t} (\cdot)$, where $\log p_{T-t}$ is estimated by target score identity \citep[TSI, ][]{de2024target} with a self-normalized importance sampler:
        \begin{align}
           {\nabla \log p_{T-t} (X_t)}& \approx \sum_{n} \frac{\ptilde (X_T^{(n)})}{ \sum_{m}  \ptilde(X_T^{(m)})}\nabla \log \ptilde (X_T^{(n)}), \quad X_T^{(n)}\sim  q_{T|t}(X_T | X_t),
        \end{align}
        $q_{T|t}(X_T | X_t)$ is the importance sampling proposal chosen as $q_{T|t}(X_T | X_t)\propto p_{t|T}(X_t | X_T)$.
        In an optimal solution, the sampling process approximates the time-reversal of a VE-SDE:
         \begin{align}\label{eq:idem_y}
            dY_t = \sqrt{2 t} dW_t, \quad  Y_0 \sim {\ptarget}.
        \end{align}
        One can re-interpret the estimator regressed in iDEM in terms of the optimal drift solving a stochastic control problem~\citep{huang2021schr}. The optimal control $f^*_{\sigma_{\mathrm{init}}}$ can be expressed in terms of the score (e.g. See Remark 3.5 in \cite{reusmooth}), for any $\sigma_{\mathrm{init}}>0$ :
        \begin{align}
             f^*_{\sigma_{\mathrm{init}}}(X_t,t) = - \nabla \log \phi_{T-t}(X_t) = -\frac{X_t}{T-t +\sigma_{\mathrm{init}}^2} +\nabla \log {p_{T-t}(X_t)},
        \end{align}
        where $\phi_t(X_t) $ is the 
     value function. 
     It can be expressed as a conditional expectation via the Feynman-Kac formula with Hopf-Cole transform \citep{hopf1950partial,cole1951quasi,fleming1989logarithmic}: \begin{align}\label{eq:opt_val_func}
            \phi_t(X_t) = \mathbb{E}_{X_T \sim q_{T|t}(X_T|X_t)}\left [\frac{\ptilde}{\mathcal{N}(0, T+ \sigma_{\mathrm{init}}^2)}(X_T)\right]. 
        \end{align}
       Note the MC Estimator of $\nabla \log \phi_{T-t}(X_t)$ (e.g. Equation \ref{eq:opt_val_func}) was used in Schr\"odinger-F\"ollmer Sampler \citep[SFS, ][]{huang2021schr} to sample from time-reversal of pinned Brownian Motion, yielding an estimator akin to the one used in iDEM.
        \item Monte Carlo Diffusion \citep[MCD,][]{doucet2022score}: unlike other neural samplers, MCD's sampling process is fixed as $\mu_t = 0, \sigma_t = 1, b_t(X_t, t) = \nabla \log \pi_t(X_t)$, where $\pi_t$ is the geometric interpolation between target and prior, i.e., $\pi_t(X_t) = p_\text{target}^{\beta_t}(X_t)p_\text{prior}^{1-\beta_t}(X_t)$.
    It can be viewed as sampling with AIS using ULA as the kernel.
    Note, that this transport is non-equilibrium, as the density of $X_t$ is not necessary $\pi_t(X_t)$.
    Therefore, MCD trains a network to approximate the time-reversal of the forward process and perform importance sampling (more precisely, AIS) to correct the bias of the non-equilibrium forward process.
        \item Controlled Monte Carlo Diffusion \citep[CMCD,][]{vargas2024transport} and Non-Equilibrium Transport Sampler \citep[NETS,][]{mate2023learning,albergo2024nets}: 
        Similar to MCD, CMCD and NETS also set $b_t(X_t, t) = \nabla \log \pi_t(X_t)$ and $\pi_t$ is the interpolation between target and prior\footnote{CMCD defines $\pi_t$ with geometric interpolation between target and prior  $\pi_t(X_t) = p_\text{target}^{\beta_t}(X_t)p_\text{prior}^{1-\beta_t}(X_t)$.
         In contrast, NETS defines 
 $\pi_t$ differently depending on the target distribution. 
 For example, with a GMM target, 
 $\pi_t$ is constructed as a GMM whose components' means and variances are linearly interpolated between the target mixture components and a Gaussian around 0. We will denote this as mode interpolation.
}.
Different from MCD where the sampling process is fixed, CMCD and NETS learn $f_\theta(\cdot, t) = \mu_t(\cdot, t)$ so that the marginal density of samples $X_t$ simulated by \Cref{eq:x} will approximate $\pi_t$.
As a special case, Liouville Flow Importance Sampler \citep[LFIS,][]{tian2024liouville} fixes $\sigma_t=0$ and learns an ODE to transport between $\pi_t$.

\end{enumerate}

\begin{table}[t]
\centering
\caption{We obtain common neural samplers by combining different underlying processes and objectives.
DDS: \citet{vargasdenoising}; DIS: \citet{berneroptimal}; DDS/DIS/PIS-LV: \citep{richterimproved}; CMCD: \citep{vargas2024transport}; NETS: \citet{albergo2024nets}; PINN: \citet{sun2024dynamical}; RDMC: \citet{huang2023monte}; iDEM: \citet{akhounditerated}; SFS: \citet{huang2021schr}; LFIS: \citet{tian2024liouville}; DGFS: \citet{zhangdiffusion}. 
*RDMC and SFS only estimate the score/optimal control function by importance sampling, and do not evolve network training.
}
\label{tab:neural_samplers}\vspace{-5pt}
\resizebox{\textwidth}{!}{%
\begin{tabular}{llccccccc}
\hline
\multicolumn{1}{c}{} & \multicolumn{1}{c}{} & \multicolumn{4}{c}{\textbf{Path measure alignment}} & \multicolumn{3}{c}{\textbf{Marginal alignment}} \\ \cmidrule(lr){3-6} \cmidrule(lr){7-9} 
\multicolumn{1}{c}{} & \multicolumn{1}{c}{\textbf{}} & \textbf{KL} & \textbf{LV} & \textbf{(S)TB} & \textbf{DB} & \textbf{PINN} & \textbf{AM} & \textbf{\makecell {Score Estimate}} \\ \hline
\multirow{2}{*}{\makecell{\textbf{Time-}\\\textbf{reversal}\\\textbf{sampler}}} & \textbf{\makecell{Reversal of \\VP/VE SDE}} & \makecell{DDS,\\ DIS} & \makecell{DDS-LV,\\DIS-LV} &  &  & PINN &  & \makecell{RDMC*,\\iDEM} \\ \cline{2-9} 
 & \textbf{\makecell{Reversal \\of PBM}} & PIS & PIS-LV & DGFS & DGFS &  &  & SFS* \\ \hline
\multicolumn{2}{l}{\textbf{Escorted transport sampler}} & CMCD & CMCD &  &  & \makecell{PINN,\\NETS,\\ LFIS} & NETS &  \\ \hline
\end{tabular}%
}
\vspace{-5pt}
\end{table}

\vspace{-5pt}
\section{Simulation-free Training with Normalizing Flow Induced SDEs}\label{sec:sim_free}\vspace{-4pt}
In this section, we propose a potential design for simulation-free training of DDS and CMCD using normalizing flows (NF)\footnote{We use NF to refer to an invertible network rather than continuous normalizing flows \citep{chen2018neural}.}.
Consider a time-dependent normalizing flow defined as $F_\theta: \mathcal{X} \times [0, T] \rightarrow \mathcal{X}.$
We denote the density of the samples drawn from the normalizing flow as $q_\theta(X_t, t)$.
A key property of NFs that enables simulation-free training is their ability to generate samples $X_t \sim q_\theta(X_t, t)$ through two distinct approaches \citep{bartosh2024neural}:
\begin{enumerate}[leftmargin=*]
    \item Drawing from the base distribution \( X_\text{base} \sim p_\text{base} \) and transforming it via \( F_\theta(X_\text{base}, t) \);
    \item Drawing an initial sample \( X_\text{base} \sim p_\text{base} \), $X_0 = F_\theta(X_\text{base}, 0)$ and evolving through an ODE:
   $
        dX_t = \partial_t F_\theta(X_\text{base}, t) dt = (\partial_t F_\theta(X_\text{base}, t)|_{X_\text{base} = F_\theta^{-1}(X_t, t)} )dt
 $. 
    For simplicity, we write $\tilde{F}_\theta(X_t, t) = \partial_t F_\theta(X_\text{base}, t)|_{X_\text{base} = F_\theta^{-1}(X_t, t)}$.
    Additionally, the following SDE will have the same marginal density as the ODE for any $\sigma\geq 0$:
    \begin{align}\label{eq:nf-sde}
        dX_t = \left(\tilde{F}_\theta(X_t, t) + \sigma_t^2 \nabla\log q_\theta(X_t, t) \right)dt + \sigma_t\sqrt{2} dW_t.
    \end{align}   
\end{enumerate}
The first approach allows us to directly generate samples along the trajectory without simulation, while the second approach allows the use of the same objective as previously described in control-based samplers.
In the following, we introduce NF-DDS, leveraging normalizing flows to achieve a simulation-free training objective.
In \Cref{appendix:nf-cmcd}, we present an alternative approach, NF-CMCD, which coincides with matching the reverse Fisher divergence between marginals in all time steps.
\par

\begin{figure}[t]
    \centering
    \begin{subfigure}{0.44\textwidth}
      \includegraphics[height=60pt, trim={0 0 80 0}, clip]{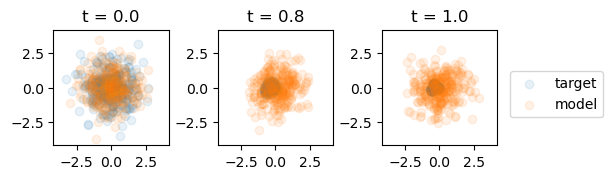}
    \caption{Initialization of NF-DDS, samples generated at different time steps $0, 0.8, 1.0$. As we can see, the initialization already covers all modes.}
    \label{fig:init_nf_dds}  
    \end{subfigure}\hfill
    \begin{subfigure}{0.55\textwidth}
      \includegraphics[height=60pt]{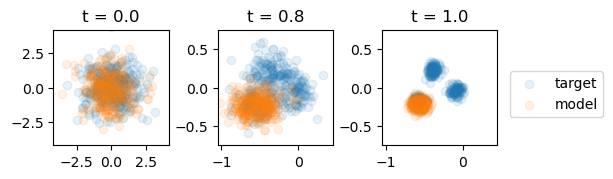}
    \caption{NF-DDS after training with \Cref{eq:flowDDS-obj}, samples generated at different time steps $0, 0.8, 1.0$.
   Unlike DDS, NF-DDS fails to capture all modes.}
    \label{fig:nf_dds}  
    \end{subfigure}\vspace{-7pt}
\end{figure}

\textbf{NF-DDS:} 
Recall that in DDS, we match the sampling process in \Cref{eq:nf-sde} with the time-reversal of a VP-SDE starting from the target density: \vspace{-2pt}
\begin{align}\label{eq:target}
   dY_t = -\beta_t Y_t dt + v \sqrt{2\beta_{t}} dW_t, \quad  Y_0 \sim \ptarget. 
\end{align}
To have a bounded RND between the target path measure and \Cref{eq:nf-sde}, we set $\sigma_t= v\sqrt{\beta_{T-t}}$. 
We rewrite the sampling process for easy reference: \vspace{-2pt}
\begin{align}\label{eq:nf-sde-dds}
        dX_t = \left(\tilde{F}_\theta(X_t, t) + v^2\beta_{T-t} \nabla\log q_\theta(X_t, t) \right)dt + v\sqrt{2\beta_{T-t}} dW_t.
    \end{align}  
By Nelson's condition \citep{nelson,ANDERSON1982313}, we can write its time-reversal as  \vspace{-2pt} 
\begin{align}\label{eq:nf-sde-dds-backward}
      dY_t = -\left(\tilde{F}_\theta(Y_t, T-t) - v^2\beta_{t} \nabla\log q_\theta(Y_t, T-t) \right)dt + v\sqrt{2\beta_{t}} dW_t, Y_0 \sim q_\theta(Y_0, T).
    \end{align} 
By Girsanov theorem, the KL divergence $D_\text{KL}[\mathbb{Q}||\mathbb{P}]$ between the path measure induced by \Cref{eq:nf-sde-dds-backward} (denoted as $\mathbb{Q}$) and \Cref{eq:target} (as $\mathbb{P}$) is tractable (derivation details in \Cref{appendix:nf-dds}): \vspace{-6pt}
\begin{align}\label{eq:flowDDS-obj}
  \!
   \! \int_0^T\!\!\! \!\frac{1}{\text{\footnotesize$4v^2\beta_{T-t}$}} \E_{{q_\theta(Y, t)}} \|
\tilde{F}_\theta(Y, t) \!- \!v^2\beta_{T-t} \nabla\log q_\theta(Y, t) \!- \!\beta_{T-t} Y
    \|^2 dt\!+ \!D_\text{KL}[q_\theta(\cdot, T)|| \ptarget].
\end{align}
\textbf{Failure of NF-DDS: } Although NF-DDS enables simulation-free training for DDS, it struggles to perform well even on simple tasks. 
We evaluate NF-DDS by training it on a 2D 3-mode Gaussian Mixture target distribution. \Cref{fig:init_nf_dds,fig:nf_dds} illustrate the initialization and the outcomes after training. 
Despite starting with an initialization that covers all modes, and being optimized using the same objective as DDS, NF-DDS fails to achieve satisfactory results.
\par
\emph{What is the difference between DDS and NF-DDS leading to this performance discrepancy}?
Excluding the influence of objectives, the only difference left is the model.
Specifically,  DDS adopts the network proposed by PIS \citep{zhangpath}:
\begin{align}\label{eq:DDS_network}
  f_\theta(\cdot, t) = \text{NN}_{1,\theta}(\cdot, t) + \text{NN}_{2, \theta}(t) \circ  \nabla\log \ptarget(\cdot),
\end{align}
and initializes $\text{NN}_{1,\theta} \approx 0$ and $\text{NN}_{2, \theta}=1$.
In the early stages of training, DDS simulation closely resembles running MCMC with Langevin dynamics and an approximation to the optimal solution (detailed in \Cref{appendix:langevin_approx}). 
In fact, nearly all algorithms discussed in \Cref{sec:review} incorporate a similar term, either explicitly or implicitly. 
If simulating these Langevin terms plays a crucial role, then modifying current algorithms to achieve simulation-free training may not be straightforward or even infeasible.
Therefore, in the next section, we provide ablation studies on the influence of the Langevin term, which we denote as \emph{Langevin preconditioning}, in both time-reversal sampler and escorted transport sampler, trained with different objectives.

\section{Ablation on Langevin Preconditioning and Its Implications}

In this section, we ablate the effectiveness of the Langevin preconditioning on examples of different neural samplers.  
For the time-reversal sampler, we take DDS as an example, while for the escorted transport sampler, we take CMCD as an example.
We will explore objectives including reverse KL, Log-var divergence, trajectory balance, and PINN.

First, we discuss {how to remove Langevin preconditioning in different samplers}:
\begin{itemize}[leftmargin=*]
    \item \emph{DDS without Langevin Preconditioning}. DDS's Langevin preconditioning occurs in its network parameterization. 
Therefore, to eliminate the help of Langevin during simulation in the training process, we can simply replace the network in \Cref{eq:DDS_network} by a standard MLP. 
To ensure the model capacity, we increase the MLP size to 5 layers with 256 hidden units.
\item \emph{CMCD without Langevin Preconditioning}. Unlike DDS, Langevin preconditioning in CMCD naturally emerges from its formulation.
Specifically, CMCD defines the drift terms for the sampling and ``target" processes as $
    f_\theta(X_t, t) +\sigma_t^2  \nabla \log \pi_t (X_t)$ and $-(f_\theta(Y_{t}, T-t) - \sigma_t^2 \nabla \log \pi_{T-t} (Y_{t}))$ respectively. By aligning their path measures, the marginal density of the sampling process at time 
$t$ is ensured to match 
$\pi_t$ in accordance with Nelson's condition \citep{nelson}.
In order to eliminate the Langevin preconditioning $\nabla \log \pi_t (X_t)$ during simulation in training, we redefine the sampling and "target" processes as $
    f_\theta(X_t, t)$ and $-(f_\theta(Y_{t}, T-t) - 2\sigma_t^2  \nabla \log \pi_{T-t} (Y_{t}))$.
Aligning their path measures still ensures that the marginal density of the sampling process at time
$t$ matches
$\pi_t$, while the training simulation does not rely on the help of Langevin preconditioning.
\item \emph{PINN without Langevin Preconditioning}. In CMCD/NETS, sampling process is defined as $
  dX_t =   \left(f_\theta(X_t, t) +\sigma_t^2  \nabla \log \pi_t (X_t)\right) dt + \sigma_t\sqrt{2} dW_t$, and the objective is independent of the value of $\sigma_t$.
Therefore, we simply set $\sigma_t = 0$ during training to eliminate the Langevin preconditioning. 
\end{itemize}
 Additionally, we investigate the performance of DDS and CMCD when the initialization is close to optimal.
To achieve this, we first train DDS and CMCD with Langevin preconditioning until convergence. 
Then, we use a new network without Langevin preconditioning to distill the teacher output with Langevin preconditioning at each time step using an $L_2$ loss.
After distillation, we fine-tune the student network using different objectives. 
This allows us to examine whether Langevin preconditioning primarily aids in localizing the model in the early training stage or also contributes to stabilizing the results in the end of training.

For DDS, we also test the results using a network conditioned on the target density instead of the target score: $f_\theta(X, t) = \text{NN}_{\theta}(X, \log \tilde{p}_\text{target}(X), t)$. 
This allows us to verify whether neural samplers require an explicit score term to ensure that the simulation behaves similarly to running Langevin dynamics, or if they only need some information about the target density.

\begin{table}[t]
\centering
\caption{Sample quality of time-reversal sampler and escorted transport sampler trained with different objectives.
We compare their performances both with and without the Langevin preconditioning. 
We measure MMD, EUBO and ELBO.
MMD can have a comprehensive reflection on the sample quality, and the difference between EUBO and ELBO measures the mode coverage: large EUBO indicates mode collapsing. 
As some methods diverge in the end, we report the results with early stopping, according to ELBO.  N/A denotes unstable training, and no reasonable result is obtained.
}
\label{tab:main_result}
\resizebox{\textwidth}{!}{%
\begin{tabular}{@{}lccccccccc@{}}
\toprule
\multirow{2}{*}{\textbf{Obj.}} & \multicolumn{5}{c}{\textbf{DDS}} & \multicolumn{4}{c}{\textbf{CMCD}} \\
\cmidrule(lr){2-6}\cmidrule(lr){7-10}  
 & \multicolumn{1}{c}{\small w. LG} & \multicolumn{1}{c}{\small w/o LG} &  
 \multicolumn{1}{c}{\small w. $\log p_\text{target}$} & \multicolumn{1}{c}{\small distil init.}  & \multicolumn{1}{c}{\small w/o LG + init.}  & \multicolumn{1}{c}{\small w. LG} & \multicolumn{1}{c}{\small w/o LG} & \multicolumn{1}{c}{\small distil init.} & \multicolumn{1}{c}{\small w/o LG + init.} \\ \midrule
\multicolumn{10}{c}{MMD ($\downarrow$)} \\ \midrule \small
rKL & 0.074 & 1.497 & 4.260 & \multirow{3}{*}{0.121} & 0.333 & 0.075 & 4.011 &   & 1.827\\
LV & 0.064 & 1.938 & 1.995 & &  0.014 &  0.017  & N/A & 0.079 & 0.036 \\
TB & 0.054 & 4.413 & 4.550 & &  0.015 &  0.035 & N/A &  & 0.130  \\
\midrule 
\multicolumn{10}{c}{\normalsize ELBO ($\uparrow$) /EUBO ($\downarrow$)} \\ \midrule \small
rKL & -0.45/0.49 & -1.93/28.52 &  -2.36/35.02 & \multirow{3}{*}{-0.88/0.64} & 
 -1.14/3.03 & -0.40/0.45  & -4.45/193.06 &  & -3.28/3$\times 10^5$  \\
LV & -0.90/0.77 & -2.07/16.26 & -1.96/17.19 & &  -0.53/0.44 & -0.28/0.33   & N/A  & -0.89/0.82  & -0.53/0.77 \\
TB & -1.73/1.36 & -2.62/23.00 & -2.61/28.75 & &  -0.46/0.45 &  -0.52/0.77   & N/A  &   & -0.77/1.20 \\
\bottomrule\normalsize
\end{tabular}%
}\vspace{-10pt}
\end{table}

\begin{figure}[t]
    \centering
    \begin{minipage}{0.43\textwidth}
        \centering
           \captionof{table}{Sample quality (MMD) by NETS trained with PINN loss \citep[][Alg 1]{albergo2024nets}, both with and without LG in the simulation process during training. As NETS used a different prior and interpolation ($\mathcal{N}(0, 2I)$, mode interpolation) compared to CMCD ($\mathcal{N}(0, 30^2I)$, geometric interpolation), we present the results by both settings for a fair investigation. N/A suggests diverging.\vspace{-6pt}}\label{tab:pinn}
            \begin{adjustbox}{width=\linewidth}
            \begin{tabular}{@{}lccc@{}}
            \toprule
            interpolant & prior & train w. LG & train w/o LG \\ \midrule
            \multirow{2}{*}{geom} & $\mathcal{N}(0, 2I)$ & 6.9529 & 7.0091 \\
            & $\mathcal{N}(0, 30^2I)$ & 0.3368 & 0.1721 \\
            \multirow{2}{*}{mode} & $\mathcal{N}(0, 2I)$ & 0.0034  & 0.0040 \\
            & $\mathcal{N}(0, 30^2I)$ & N/A & N/A  \\ \bottomrule
            \end{tabular}
            \end{adjustbox}\vspace{0pt}
    \end{minipage}\hfill
    \begin{minipage}{0.55\textwidth}
        \centering
        \includegraphics[width=\linewidth]{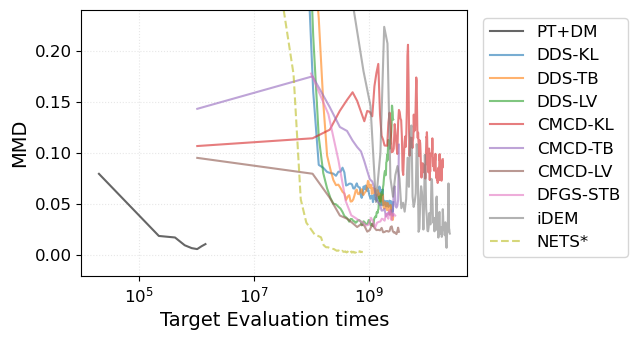}
        \caption{Sample quality vs target evaluation times for different approaches with different objectives on GMM-40 target.  *NETS uses mode interpolation, which is distinct from that employed in others. }\label{fig:energy_call_mmd}\vspace{0pt}
    \end{minipage}\\
     \begin{minipage}{\textwidth}
        \centering
        \includegraphics[width=\linewidth, trim={80 150 100 100}, clip]{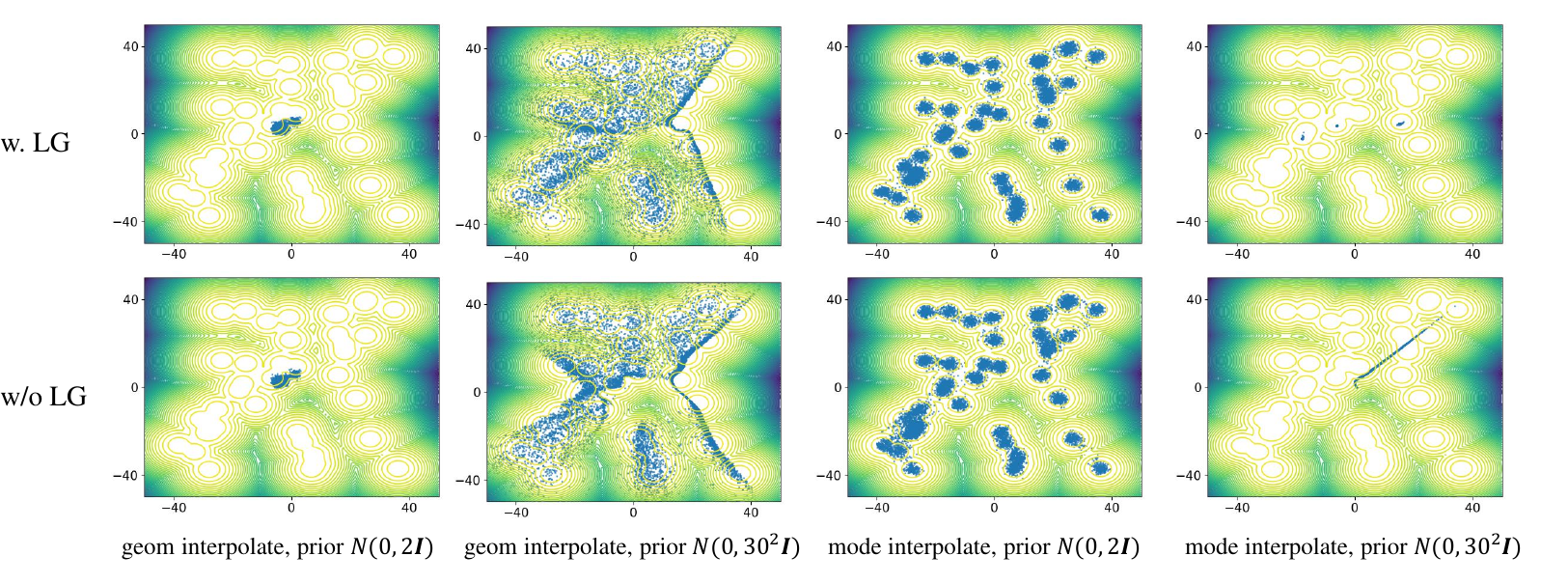}
        \caption{Sampled obtained by PINN with different settings.
   We can see PINN seems to be highly robust to Langevin preconditioning. 
   However, it is highly sensitive to the prior and interpolation. }\label{fig:pinn_sample}\vspace{0pt}
    \end{minipage}%
\end{figure}

We present results for DDS and CMCD using reverse KL (rKL), log-variance divergence (LV), and trajectory balance (TB) on a 40-mixture Gaussian target proposed by \citet{midgleyflow} in \Cref{tab:main_result}, with more visualization in \Cref{appendix:visualize}.
Our findings reveal the following key observations:
\begin{itemize}[leftmargin=*]
    \item \textbf{Most objectives significantly collapse without Langevin preconditioning.}
We note that, at initialization, the samples from the neural samplers already cover all modes, meaning there is no inherent exploration issue. 
However, even with this favorable initialization, the absence of Langevin preconditioning leads to severe collapse in most objectives.
\item \textbf{Langevin preconditioning cannot be replaced by alternative target information, such as} $\log p_\text{target}$.
This suggests that neural samplers require an explicit score term to ensure that the simulation behaves similarly to Langevin dynamics.
\item \textbf{If the initialization is close to optimal, TB and LV refine the solution more stably, while rKL remains prone to mode collapse.}
This suggests that future work could explore a training pipeline where the sampler is first warmed up using Langevin dynamics, followed by fine-tuning with these objectives to reduce the number of target energy evaluations during sampling.
\end{itemize}
We also include results obtained by NETS with the PINN loss in \Cref{tab:pinn}  with more visualization in \Cref{fig:pinn_sample}.
Since NETS employs a different prior and interpolation scheme compared to CMCD in \Cref{tab:main_result}, we present results for both settings to ensure a fair comparison.
Surprisingly, we observe that \textbf{the PINN loss is relatively robust to Langevin preconditioning during simulation}.
Additionally, by design, the PINN loss naturally supports simulation-free training.
However, its performance is highly sensitive to the choice of prior and interpolation:
a large prior leads to diverging in mode interpolation, while a smaller one also fails under geometric interpolation.
Furthermore, the PINN loss requires computing an expensive divergence term, making it challenging to apply to simulation-free approaches with normalizing flows proposed in \Cref{sec:sim_free}.

Finally, the critical role of Langevin preconditioning naturally raises an important question:
Is simulation during training with Langevin preconditioning more efficient than directly generating data with Langevin dynamics and fitting a model post hoc?
Unfortunately, the answer is no.
In \Cref{fig:energy_call_mmd}, we compare several neural samplers against an alternative approach where Parallel Tempering (PT) \citep[PT, ][]{PhysRevLett.57.2607,earl2005parallel} is first used to generate samples, followed by fitting a diffusion model. 
We assess both sample quality and the number of target energy evaluations required.
The results clearly show that \textbf{almost all neural samplers require several orders of magnitude more target evaluations compared to PT.}

\section{Discussions and Conclusions}
Motivated by the pursuit of simulation-free training, we reviewed neural samplers from the perspective of sampling processes and training objectives, as well as revisiting their dependence on Langevin preconditioning.
Our findings reveal that most training methods for diffusion and control-based neural samplers heavily rely on Langevin preconditioning.
While PINN appears to be an exception, it still requires evaluating both the target density and the model’s divergence at every time step along the trajectory, making it no more efficient in practice.
This highlighted critical caveats in scaling neural samplers to high-dimensional and real-world problems.
In fact, while significant advances have been made in learning neural samplers directly from unnormalized densities, the most efficient and practical approach remains running MCMC first and fitting a generative model post hoc.
\par
Our results reveal several open questions and future directions worth exploring.
First, talking about neural samplers, many works focus on learning models directly from the unnormalized density, avoiding the use of any data from the target density. 
However, given that the Langevin preconditioning plays a crucial role in most approaches, we may equivalently interpret the training process as running several steps of MCMC to obtain approximate samples.
This interpretation, blurring the distinction between data-driven and data-free approaches,  challenges the definition of these ``data-free" neural samplers.
 Furthermore, as our results demonstrate, a straightforward two-step approach---first running Parallel Tempering (PT) to obtain samples, followed by fitting a diffusion model---yields significantly higher efficiency compared to nearly all neural samplers.
This observation further questions the practical justification and motivation of ``data-free" neural samplers.
Therefore, rather than attempting to completely avoid the use of data, \textbf{a more promising and practical direction may involve developing objective functions or training pipelines that rely on a limited amount of data for a more efficient acquisition of information from each target density evaluation.}
\par
However, we emphasize that while we advocate for the explicit utilization of data, we acknowledge that \textbf{it may not always be feasible, or even reasonable, for newly developed approaches to surpass these well-established baselines from the outset.}
The methods developed within ``data-free" training pipelines remain valuable and can provide inspiration for approaches that more effectively leverage data, potentially leading to improved efficiency and performance in neural samplers.
\par
Based on our observations, PINN loss appears to be an example with such potential. 
It demonstrates greater robustness in the absence of Langevin preconditioning and naturally supports simulation-free training by its design. 
However, it still requires extensive target evaluations along the entire trajectory and tends to be more sensitive to hyperparameters, making it challenging to apply as an off-the-shelf method.
Therefore, \textbf{future research could focus on learning better priors or interpolations.} 
A straightforward approach may involve first obtaining approximate samples from the target distribution using methods such as MCMC, then learning priors or interpolants from these samples, and using the learned hyperparameters to refine the sample quality in an iterative manner.

\section*{Acknowledgment}
We thank Julius Berner, Brian Lee, Ezra Erives and Bowen Jing for helpful suggestions regarding implementations of several neural samplers and objective functions. We thank Michael Albergo for aiding us in running the NETS and PINN based experiments. 
JH and SP acknowledge support from the University of Cambridge Harding Distinguished Postgraduate Scholars Programme.  
JMHL and RKOY acknowledge support from a Turing AI Fellowship under grant EP/V023756/1. 
CPG and YD acknowledge the support of Schmidt Sciences programs, an AI2050 Senior Fellowship and Eric and Wendy Schmidt AI in Science Postdoctoral Fellowships; the National Science Foundation (NSF); the National Institute of Food and Agriculture (USDA/NIFA); the Air Force Office of Scientific Research (AFOSR).

\newpage
\bibliography{iclr2025_conference}
\bibliographystyle{iclr2025_conference}

\newpage
\appendix

\section{Taxonomy of Objective Functions}

In this section, we briefly describe different objectives that we reviewed and used in the main text.

\subsection{Path measure alignment objectives}

The \textit{path measure alignment} framework aims to align the 
sampling process starting from $p_\text{prior}$ to a      ``target"  process starting from $p_\text{target}$. 
In the following, we denote  $\mathbb{Q}$ as the sampling process  and $\mathbb{P}$ as the ``target" process.
However, we should note that this notation does not necessarily imply that $\mathbb{Q}$ is the process parameterized by the model. 
In fact, this is only true for samplers like PIS or DDS. 
For escorted transport samplers like CMCD, both  $\mathbb{Q}$ and  $\mathbb{P}$ involve the model, and for annealed variance reduction sampler like MCD,  $\mathbb{Q}$ is fixed, and the model only appears in $\mathbb{P}$.
We now describe five commonly used objectives:

\textbf{Reverse KL divergence.} 
Reverse KL divergence is defined as
\begin{align}
    D_\mathrm{KL}[\mathbb{Q}||\mathbb{P}] = \E_{\mathbb{Q}} \left[
  \log   \frac{\mathrm{d} \mathbb{Q}}{\mathrm{d} \mathbb{P}}
    \right].
\end{align}
In practice, we approximate the expectation with Monte Carlo estimators, and calculate the log Radon–Nikodym derivative $\log   \frac{\mathrm{d} \mathbb{Q}}{\mathrm{d} \mathbb{P}}$, either through Gaussian approximation via Euler–Maruyama discretization or by applying Girsanov's theorem.

\textbf{Log-variance divergence.} Log-variance divergence optimizes the second moment of the log ratio
\begin{align}
    D_\mathrm{logvar}[\mathbb{Q}||\mathbb{P}] = \text{Var}_{\tilde{\mathbb{Q}}} \left(
  \log   \frac{\mathrm{d} \mathbb{Q}}{\mathrm{d} \mathbb{P}}
    \right).
\end{align}
Unlike KL divergence, which requires the expectation to be taken with respect to \(\mathbb{Q}\), log-variance allows the variance to be computed under a different measure \(\tilde{\mathbb{Q}}\). 
This flexibility suggests that we can detach the gradient of the trajectory or utilize a buffer to stabilize training.
On the other hand, when the variance is taken under  \(\mathbb{Q}\), the gradient of log-variance divergence w.r.t parameters in  \(\mathbb{Q}\) is the same as that of reverse KL divergence~\citep{richter2020vargrad}:
\begin{align}
    \frac{\mathrm{d}}{\mathrm{d} \mathbb{\theta}} \text{Var}_{\tilde{\mathbb{Q}}} \left(
  \log   \frac{\mathrm{d} \mathbb{Q}_{\theta}}{\mathrm{d} \mathbb{P}}
    \right)\Bigg|_{\tilde{\mathbb{Q}} = {\mathbb{Q}_{\theta}}} =    \frac{\mathrm{d}}{\mathrm{d} {\theta}}  D_\mathrm{KL}[\mathbb{Q}_{\theta}||\mathbb{P}].
\end{align}
However, we note that this conclusion holds only \emph{in expectation}.
In practice, when the objective is calculated with Monte Carlo estimators, they will exhibit different behavior.

\textbf{Trajectory balance.} Trajectory balance optimizes the squared log ratio
\begin{align}
    D_\mathrm{TB}[\mathbb{Q}||\mathbb{P}] = \E_{\tilde{\mathbb{Q}}} \left[
  \left(\log   \frac{\mathrm{d} \mathbb{Q}}{\mathrm{d} \mathbb{P}} - k\right)^2
    \right],
\end{align}
which is equivalent to the log-variance divergence with a learned baseline $k$.

\textbf{Sub-trajectory balance.} 
TB loss matches the entire $\mathbb{Q}$ and $\mathbb{P}$ as a whole. 
Alternatively, we can match segments of each trajectory individually to ensure consistency across the entire trajectory. This approach leads to the sub-trajectory balance objective. 
For simplicity, though it is possible to define sub-trajectory balance in continuous time, we define it with time discretization.
\par
With Euler–Maruyama discretization, we discretize $\mathbb{Q}$ and $\mathbb{P}$ into sequential produce of measure, with density given by:
\begin{align}
 p_0(X_0) \prod_{n=0}^{N-1} p_F(X_{n+1}|X_n) \quad\text{and}\quad   \ptilde(X_N) \prod_{t=0}^{N-1} p_B(X_n|X_{n+1}).
\end{align}
Note that the density for discretized $\mathbb{P}$ can be unnormalized.

Then, we introduce a sequence of intermediate densities \(\{\pi_n\}_{n=0}^{N}\), where the boundary conditions are given by \(\pi_0 = p_{\text{prior}}\) and \(\pi_N = \tilde{p}_{\text{target}}\). 
These intermediate distributions can either be prescribed as a fixed interpolation between the target and prior distributions or be learned adaptively through a parameterized neural network.

Finally, we define the sub-trajectory balance objective as
\begin{align}
      D_\mathrm{STB}[\mathbb{Q}||\mathbb{P}] = \E_{\tilde{\mathbb{Q}}} \left[ \sum_{0 \leq i<j\leq N}
  \left(\log   \frac{\pi_i(x_i) \prod_{n=i}^{j-1} p_F(x_{n+1}|x_n) }{\pi_j(x_j) \prod_{n'=i}^{j-1} p_B(x_{n'}|x_{n'+1})} + k_i - k_j\right)^2
    \right].
\end{align}

\textbf{Detailed balance.} 
Detailed balance can be viewed as an extreme case of sub-trajectory balance, where instead of summing over sub-trajectories of all lengths, we only calculate the sub-trajectory balance over each discretization step:

\begin{align}
      D_\mathrm{DB}[\mathbb{Q}||\mathbb{P}] = \E_{\tilde{\mathbb{Q}}} \left[ \sum_{0 \leq i\leq N-1}
  \left(\log   \frac{\pi_i(x_i) p_F(x_{i+1}|x_i) }{\pi_j(x_{i+1})  p_B(x_{i}|x_{i+1})} + k_i - k_{i+1}\right)^2
    \right].
\end{align}

\subsection{Marginal alignment objectives}
Unlike path measure alignment, \emph{marginal alignment} objectives directly enforce the sampling process at each time step $t$ to match with some marginal $\pi_t$.
$\pi_t$ can be either prescribed as an interpolation between the target and prior, with boundary conditions $\pi_0 = p_\text{prior}$ and $\pi_T = p_\text{target}$, or be learned through a network under the constraint of the boundary conditions. 
Commonly used objectives in this framework include PINN and action matching:

\textbf{PINN.} For the sampling process defined by $ dX_t =   \left(f_\theta(X_t, t) +\sigma_t^2  \nabla \log \pi_t (X_t)\right) dt + \sigma_t\sqrt{2} dW_t$, the PINN loss is given by
\begin{align}
    \mathcal{L}_{\text{PINN}} = \int_0^T \mathbb{E}_{\tilde{q}_t(X_t)}  ||
  \nabla \cdot f_\theta(X_t, t) + \nabla \log\pi_t(X_t) \cdot f_\theta(X_t, t) + (\partial_t \log\pi_t)(X_t) + \partial_t F(t) 
    ||^2 dt,
\end{align}
where $F(t) $ is parameterized by a neural network. 
Note that the expectation can be taken over an arbitrary $\tilde{q}_t$, as long as the marginal of $\mathbb{Q}$ at time $t$ is absolute continuous to $\tilde{q}_t$.
We also note PINN does not depend on the specific value of $\sigma_t$ in the sampling process.

\textbf{Action matching.} Similar to PINN, an action matching-based~\citep{neklyudov2023action} objective is derived by~\citep{albergo2024nets} for the PDE-constrained optimization problem
\begin{multline}
\mathcal{L}_{\text{AM}} =  \int_0^T  \mathbb{E}_{q_t(X_t)} \left[ \frac{1}{2} ||\nabla \phi_t(X_t)||^2 + \partial_t \phi_t(X_t)\right] dt \\+ \mathbb{E}_{p_{\text{prior}}(X_0)} \left[ \phi_0(X_0)\right] - \mathbb{E}_{p_{\text{target}}(X_T)} \left[ \phi_T(X_T) \right] ,
\end{multline}
where the vector field $b_t = \nabla \phi_t$, induced by a scalar potential, and $\phi_t$ is called the ``action''.

\section{Detailed Summary of Samplers}\label{appendix:review}
In this section, we provide a more detailed review of diffusion and control-based neural samplers.
We also discuss how these neural samplers rely on the Langevin preconditioning in the end.
\subsection{Sampling process and Objectives}
We write the sampling process as follows:
    \begin{align}\label{eq:x_appendix}
        d X_t = \left [\mu_t(X_t) + \sigma_t^2 b_t(X_t) \right] dt + \sigma_t \sqrt{2} dW_t, \quad X_0 \sim p_\text{prior},
    \end{align}

 \begin{enumerate}[label=({{\arabic*}}), leftmargin=*]
        \item Path Integral Sampler \citep[PIS,][]{zhangpath} and concurrently \citep[NSFS, ][]{vargas2021bayesian}: 
        PIS fixes $p_\text{prior} = \delta_0,  \sigma_t=1/\sqrt{2}$ and learns a network $f_\theta(\cdot) =\mu_t(\cdot) + \sigma_t^2 b_t(\cdot)$ so that \Cref{eq:x_appendix} approximate the time-reversal of the following SDE (Pinned Brownian Motion):
        \begin{align}\label{eq:pis_y_appendix}
             d Y_t = -\frac{Y_t}{T-t} dt + dW_t, \quad Y_0 \sim \ptarget.
        \end{align}
        We define \Cref{eq:pis_y_appendix} as the time-reversal of \Cref{eq:x_appendix} when $Y_t \sim X_{T-t}$.
        The network is learned by matching the reverse KL \citep{zhangpath,vargas2021bayesian} or log-variance divergence \citep{richterimproved} between the path measures of the sampling and the target process.
       
       \item Diffusion generative flow samplers~\citep[DFGS,][]{zhangdiffusion} learns to sample from  the same process as PIS,  but with a new introduction of local objectives including detailed balance and (sub-)trajectory balance.
        In fact, trajectory balance can been shown to be equivalent to the log-variance objective with a learned baseline rather than a Monte Carlo (MC) estimator for the first moment~\citep{nusken2021solving}.  
        \item Denoising Diffusion Sampler \citep[DDS,][]{vargasdenoising} and time-reversed Diffusion Sampler \citep[DIS,][]{berneroptimal}: 
        both DDS and DIS fix $\mu_t(X_t, t) = \beta_{T-t} X_t, \sigma_t = v \sqrt{\beta_{T-t}}, p_\text{prior} = \mathcal{N}(0, v^2I)$, and learn a network $f_\theta(\cdot, t) = b_t(\cdot, t)/2 $ so that \Cref{eq:x_appendix} approximates the time-reversal of the VP-SDE:
        \begin{align}\label{eq:dds_y_appendix}
            dY_t = -\beta_t Y_t dt + v \sqrt{2\beta_{t}} dW_t, \quad  Y_0 \sim \ptarget.
        \end{align}
        Similar to PIS, the network can be trained either with reverse KL divergence or log-variance divergence.
        In an optimal solution, $f_\theta$ will approximate the score $f_\theta(\cdot, t) \approx \nabla\log p_{T-t}(\cdot)$, where $ p_{t}(X) = \int  \mathcal{N}(X|\sqrt{1-\lambda_t}Y, v^2\lambda_t I) \ptarget(Y)dY$ and $\lambda_t = 1-\exp(-2\int_0^t  \beta_s ds)$.
        
       \item Iterated Denoising Energy Matching \citep[iDEM,][]{akhounditerated}: iDEM fixes $\mu_t(X_t, t) = 0, p_\text{prior} = \mathcal{N}(0, T^2 I)$, and learns a network $f_\theta(\cdot, t) = b_t(\cdot, t) / 2$ to approximate the score  $f_\theta(X_t, t) \approx \nabla \log p_{T-t} (X_t)$.
        This is achieved by writing the score with target score identity \citep[TSI, ][]{de2024target}, and estimating it with a self-normalized importance sampler:
        \begin{align}
            \nabla \log p_{T-t} (X_t) &\stackrel{\text{TSI}}{=} \int p_{T|T}(X_T | X_t)  \nabla \log \ptilde (X_T) dX_T\\
            &\stackrel{\text{Bayes' Rule}}{=}\int \frac{\ptilde (X_T)p_{t|T}(X_t|X_T)}{\int \ptilde (X_T)p_{t|T}(X_t|X_T) dX_T}  \nabla \log \ptilde (X_T) dX_T\\
            &=\int q_{T|t}(X_T | X_t)\frac{\ptilde (X_T)p_{t|T}(X_t|X_T) \nabla \log \ptilde (X_T) }{q_{T|t}(X_T | X_t) \int q_{T|t}(X_T | X_t) \frac{\ptilde (X_T)p_{t|T}(X_t|X_T) }{q_{T|t}(X_T | X_t)}dX_T} dX_T.
        \end{align}
        By choosing $q_{T|t}(X_T | X_t)\propto p_{t|T}(X_t|X_T)$, we obtain
        \begin{align}
            \nabla \log p_{T-t} (X_t) &=\int q(X_T | X_t)\frac{\ptilde (X_T)\nabla \log \ptilde (X_T)}{ \int q(X_T | X_t) \ptilde (X_T)dX_T}   dX_T\\
            &\approx\sum_{n} \frac{\ptilde (X_T^{(n)})}{ \sum_{n}  \ptilde (X_T^{(n)})}\nabla \log \ptilde (X_T^{(n)}), \quad X_T^{(n)}\sim  q_{T|t}(X_T | X_t)\\
            & =: \widehat{\nabla \log p_{T-t} (X_t)}.
        \end{align}
        Then, iDEM matches $f_\theta(X_t, t)$ with $\widehat{\nabla \log p_{T-t} (X_t)}$ by $L^2$ loss.
        In optimal, the sampling process will approximate the time-reversal of a VE-SDE:
         \begin{align}\label{eq:idem_y_appendix}
            dY_t =\sqrt{2t} dW_t, \quad  Y_0 \sim p_\text{target}.
        \end{align}
Several extensions have been developed based on iDEM:
Bootstrapped Noised Energy Matching \citep[BNEM, ][]{ouyang2024bnemboltzmannsamplerbased} generalizes the self-normalized importance sampling estimator of the score to the energy function, enabling the training of energy-parameterized diffusion models.
They also proposed a bootstrapping approach to reduce the training variance.
Diffusive KL \citep[DiKL, ][]{he2024training} integrates this estimator with variational score distillation techniques \citep{poole2022dreamfusion, luo2024diff} to train a one-step generator as the neural sampler.
Also, DiKL proposes using MCMC to draw samples from $p_{T|t}(X_T|X_t)$ to estimate the score,  instead of relying on the self-normalized importance sampling estimator with the proposal $q_{T|t}(X_T | X_t)\propto p_{t|T}(X_t|X_T)$, leading to lower variance during training.
\par
~ We also note that iDEM's score estimator is closely related to stochastic control problems.
   One can re-express the estimator regressed in iDEM in terms of the optimal drift of a stochastic control problem~\citep{huang2021schr}, the optimal control $f^*$ can be expressed in terms of the score (e.g. See Remark 3.5 in \cite{reusmooth}) :
        \begin{align}\label{eq:optimal_drift}
             f^*_{t}(X_t) = - \nabla \log \phi_{T-t}(X_t) = - \nabla \ln \nu_{T-t}^{\mathrm{ref}}(X_T) +\nabla \log {p_{T-t}(X_t)},
        \end{align} where $\phi_t(X_t) $ is the 
     value function, which can be expressed as a conditional expectation via the Feynman-Kac formula followed by the Hopf-Cole transform \citep{hopf1950partial,cole1951quasi,fleming1989logarithmic}:
\begin{align}\label{eq:opt_val_func_appendix}
\phi_t(x) = \mathbb{E}_{X_T \sim q_{T|t}(X_T|x)}\left [\frac{\ptilde}{\nu^{\mathrm{ref}}_T}(X_T)\right]. 
\end{align}
Where in the case for VE-SDE (i.e. iDEM) and 
$\nu^{\mathrm{ref}}_t(x) = \mathcal{N}(x| 0, t+ \sigma_{\mathrm{prior}}^2)$ and thus $\phi_{T-t}(X_t) = \frac{X_t}{T-t +\sigma_{\mathrm{init}}^2} +\nabla \log {p_{T-t}(X_t)}$.

Note the MC Estimator of $\nabla \log \phi_{T-t}(X_t)$ (e.g. Equation \ref{eq:opt_val_func}) was used in Schr\"odinger-F\"ollmer Sampler \citep[SFS, ][]{huang2021schr} to sample from time-reversal of pinned Brownian Motion, yielding an akin estimator to the one used in iDEM, in particular they carry out an MC estimator of the following quantity:

\begin{align}
   \nabla \phi_{T-t}(x) = \frac{\mathbb{E}_{Z \sim \mathcal{N}(0,I)}\left [\nabla\frac{\ptilde}{\nu^{\mathrm{ref}}_T}( \mu_{T|T-t}x + \sigma_{T|T-t}Z)\right]}{\mathbb{E}_{Z \sim \mathcal{N}(0,I)}\left [\frac{\ptilde}{\nu^{\mathrm{ref}}_T}( \mu_{T|T-t}x + \sigma_{T|T-t}Z)\right]}.
\end{align}

Where, we have assumed that $q_{T|t}(x_T|x_t) =\mathcal{N}(x_T| \mu_{T|t}x_t, \sigma_{T|t})$ as is the case with most time reversal based samplers and generative models.

        \item Monte Carlo Diffusion \citep[MCD,][]{doucet2022score}: unlike other neural samplers, MCD's sampling process is fixed as $\mu_t = 0, \sigma_t = 1, b_t(X_t, t) = \nabla \log \pi_t(X_t)$, where $\pi_t$ is the geometric interpolation between target and prior, i.e., $\pi_t(X_t) = p_\text{target}^{\beta_t}(X_t)p_\text{prior}^{1-\beta_t}(X_t)$.
    It can be viewed as sampling with AIS using ULA as the kernel.
    Note, that this transport is non-equilibrium, as the density of $X_t$ is not necessary $\pi_t(X_t)$.
    Therefore, MCD trains a network to approximate the time-reversal of the forward process and perform importance sampling (more precisely, AIS) to correct the bias of the non-equilibrium forward process.
        \item Controlled Monte Carlo Diffusion \citep[CMCD,][]{vargas2024transport} and Non-Equilibrium Transport Sampler \citep[NETS,][]{mate2023learning,albergo2024nets}: 
        Similar to MCD, CMCD and NETS also set $b_t(X_t, t) = \nabla \log \pi_t(X_t)$ and $\pi_t$ is the interpolation between target and prior.
Different from MCD where the sampling process is fixed, CMCD and NETS learn $f_\theta(\cdot, t) = \mu_t(\cdot, t)$ so that the marginal density of samples $X_t$ simulated by \Cref{eq:x} will approximate $\pi_t$.
As a special case, Liouville Flow Importance Sampler \citep[LFIS,][]{tian2024liouville} fixes $\sigma_t=0$ and learns an ODE to transport between $\pi_t$.

\end{enumerate}

\subsection{Lanvegin Preconditioning in Diffusion/Control-based Neural Samplers}
\label{appendix:langevin_precond}

\textbf{Explicit Langevin preconditioning. }
In samplers including PIS, DDS, DFGS, DIS, etc., 
The network is parameterized with a skip connection using Lanvegin preconditioning:
\begin{align}
  f_\theta(\cdot, t) = \text{NN}_{1,\theta}(\cdot, t) + \text{NN}_{2, \theta}(t) \circ  \nabla\log \ptarget(\cdot).
\end{align}
In samplers like MCD, the forward process is a sequence of Lanvegin dynamics with invariant density $\pi_t$ as the interpolation between prior and target.
In CMCD and NETS, the drift of forward process is given by the network output plus a score term: 
\begin{align}
    f_\theta(X_t, t) +\sigma_t^2  \nabla \log \pi_t (X_t).
\end{align}
\textbf{Implicit Langevin preconditioning. } In iDEM, we regress the network with 
\begin{align}
     \nabla \log p_{T-t} (X_t) 
            &\approx\sum_{n} \frac{\ptilde (X_T^{(n)})}{ \sum_{n}  \ptilde (X_T^{(n)})}\nabla \log \ptilde (X_T^{(n)}), \quad X_T^{(n)}\sim  q_{T|t}(X_T | X_t).
\end{align}
Note that while the network does not explicitly depend on the score of the target density, the objective compels it to learn gradient information.
This gradient information is utilized during simulation when collecting the buffer every few iterations, effectively inducing an implicit Langevin preconditioning.

\textbf{No Langevin preconditioning. } 
LFIS does not rely on Langevin preconditioning during simulation.
Like NETS, it employs the PINN loss, but its sampling process is governed by an ODE.
Thus, similar to our discussion in the main text on eliminating Langevin preconditioning for PINN-based CMCD, LFIS inherently removes this dependency in its design.

LFIS adopts several tricks to stabilize the training and ensure mode covering:
it learns the ODE drift progressively, starting from the prior and gradually transitioning to the target.
Additionally, it employs separate networks for different time steps to prevent forgetting.
But even without these tricks,  our results in \Cref{tab:pinn} confirm the robustness of the PINN loss to Langevin preconditioning when the interpolation and prior are carefully tuned.

\section{NF-DDS}\label{appendix:nf-dds}

Here we present a derivation of the NF-DDS objective.
\begin{align}
   \nonumber & D_\text{KL}[\mathbb{Q}||\mathbb{P}]\\  \nonumber = &\E_{\mathbb{Q}} \left[
    \int_0^T \frac{1}{4v^2\beta_t}\|
   \tilde{F}_\theta(Y_t, T-t) - v^2\beta_{t} \nabla\log q_\theta(Y_t, T-t) - \beta_t Y_t
    \|^2 dt
    \right] + D_\text{KL}[q_\theta(\cdot, T)|| p_\text{target}]\\ \nonumber 
   = & 
    \int_0^T \E_{\mathbb{Q}} \left[\frac{1}{4v^2\beta_t}\|
   \tilde{F}_\theta(Y_t, T-t) - v^2\beta_{t} \nabla\log q_\theta(Y_t, T-t) - \beta_t Y_t
    \|^2\right] dt+ D_\text{KL}[q_\theta(\cdot, T)|| p_\text{target}]\\ \nonumber 
   = & 
    \int_0^T \frac{1}{4v^2\beta_t}\E_{{q_\theta(Y, T-t)}}\|
   \tilde{F}_\theta(Y, T-t) - v^2\beta_{t} \nabla\log q_\theta(Y, T-t) - \beta_t Y
    \|^2 dt+ D_\text{KL}[q_\theta(\cdot, T)|| p_\text{target}]\\
  = &
    \int_0^T \frac{1}{4v^2\beta_{T-t}} \E_{{q_\theta(Y, t)}} \|
   \tilde{F}_\theta(Y, t) - v^2\beta_{T-t} \nabla\log q_\theta(Y, t) - \beta_{T-t} Y
    \|^2 dt+ D_\text{KL}[q_\theta(\cdot, T)|| p_\text{target}].\label{eq:flowDDS-obj_appendix}
\end{align}

\section{NF-CMCD}\label{appendix:nf-cmcd}
In this section, we proposed a CMCD variation with normalizing flow for simulation-free training.
\par
In CMCD, we match the forward sampling process:
\begin{align}\label{eq:nf-sde-cmcd}
        dX_t = \left(\tilde{F}_\theta(X_t, t) + \sigma_t^2 \nabla\log q_\theta(X_t, t) \right)dt + \sigma_t\sqrt{2} dW_t, X_0 \sim q_\theta(X_0, 0),
    \end{align}
 with a target backward process, calculated by Nelson's condition assuming the marginal of the SDE at each time step matches with a prescribed marginal density  e.g. $\pi_t(\cdot) = p_\text{target}^{\beta_t}(\cdot)p_\text{prior}^{1-\beta_t}(\cdot)$:
 \begin{multline}\label{eq:nf-sde-cmcd-target}
        dY_t = -\Big(\tilde{F}_\theta(Y_t, T-t) + \sigma_{T-t}^2 \nabla\log q_\theta(Y_t, T-t)\\- 2 \sigma_{T-t}^2 \nabla\log \pi_{T-t}(Y_t, T-t) \Big)dt + \sigma_{T-t}\sqrt{2} dW_t, Y_0 \sim p_\text{target}.
 \end{multline}
Again, similar to NF-DDS, the time-reversal of \Cref{eq:nf-sde-cmcd} can be calculated by Nelson's condition:
\begin{align}\label{eq:nf-sde-cmcd-backward}
     dY_t = -\left(\tilde{F}_\theta(Y_t, T-t) - \sigma_{T-t}^2 \nabla\log q_\theta(Y_t, T-t) \right)dt + \sigma_{T-t}\sqrt{2} dW_t, Y_0 \sim q_\theta(Y_0, T).
\end{align}
By Girsanov theorem, the KL divergence between the path measure by \Cref{eq:nf-sde-cmcd-backward} (denoted as $\mathbb{Q}$) and \Cref{eq:nf-sde-cmcd-target} (as $\mathbb{P}$) is:
\begin{align}
  D_\text{KL}[\mathbb{Q}||\mathbb{P}] = 
    \int_0^T \frac{1}{\sigma_{t}} \E_{{q_\theta(Y, t)}} \| \sigma_{t}^2 \nabla\log q_\theta(Y, t) - \sigma_{t}^2 \nabla\log \pi_{t}(Y, t) 
    \|^2 dt+ D_\text{KL}[q_\theta(\cdot, T)|| p_\text{target}].
\end{align}
This coincides with the Fisher divergence between each marginal.

\par
After training, we can sample from 
\begin{align}\label{eq:nf-sde-cmcd_forward}
        dX_t = \left(\tilde{F}_\theta(X_t, t) + \sigma_t^2 \nabla\log \pi_t(X_t) \right)dt + \sigma_t\sqrt{2} dW_t, X_0 \sim q_\theta(X_0, 0),
\end{align}
with approximate reversal
\begin{align}\label{eq:nf-sde-cmcd-backward_sample}
     dY_t = -\left(\tilde{F}_\theta(Y_t, T-t) - \sigma_{T-t}^2 \nabla\log \pi_{T-1}(Y_t) \right)dt + \sigma_{T-t}\sqrt{2} dW_t, Y_0 \sim \ptilde.
\end{align}

\section{Langevin Preconditioning as an Approximation to the Optimal Control}
\label{appendix:langevin_approx}

As we discussed for iDEM (similarly for PIS and DDS), the optimal drift can be expressed as the gradient of the value function in \Cref{appendix:langevin_precond}
$$f^*_{t}(X_t) = - \nabla \log \phi_{T-t}(X_t)$$ 
where the value function itself can be expressed as
\begin{align}
   \log \phi_{t}(x) = \log \mathbb{E}_{X_T \sim q_{T|t}(X_T|x)}\left [\frac{\ptilde}{\nu^{\mathrm{ref}}_T}(X_T)\right].
\end{align}
where we define the reference process to be the uncontrolled process starting from $p_\text{prior}$, $\nu^{\mathrm{ref}}_T$ is the marginal density for the last time step of the reference process, and $q_{T|t}(X_T|x)$ is the conditional density for the reference process.
Specifically, for PIS, the reference is 
\begin{align}
    d X_t = dW_t, X_0 = 0;
\end{align}
and for DDS, the reference is 
\begin{align}
     dX_t = -\beta_{T-t} X_t dt + v \sqrt{2\beta_{T-t}} dW_t, \quad  X_0 \sim p_\text{prior} = \mathcal{N}(0, v^2).
\end{align}
We hence have
\begin{align}
   & \nu^{\mathrm{ref}}_T = \mathcal{N}(0, TI) \text{ for PIS}, \quad  \nu^{\mathrm{ref}}_T = \mathcal{N}(0, v^2I)\text{ for DDS}\\
    &  \mathbb{E}[X_T |X_t=x]  = x \text{ for PIS}, \quad  \mathbb{E}[X_T |X_t=x] = \sqrt{1-\lambda_t} \text{ for DDS}
\end{align}
Now consider approximating the above expectation with a point estimate using the posterior mean (e.g. reversing Jensens Inequality):
\begin{align}
    \log \phi_{t}(x) &\approx   \log  \frac{\ptilde}{\nu^{\mathrm{ref}}_T}(\mathbb{E}[X_T |X_t=x]) \\
    &\approx  \log \ptilde (\mathbb{E}[X_T |X_t=x]) -\log \nu^{\mathrm{ref}}_T (\mathbb{E}[X_T |X_t=x])
\end{align}
Taking gradients on both sides:
\begin{align}
    \nabla_x     \log \phi_{t}(x) &\approx \nabla_x   \log \ptilde (\mathbb{E}[X_T |X_t=x]) - \nabla_x    \log\nu^{\mathrm{ref}}_T (\mathbb{E}[X_T |X_t=x])
\end{align}

where for PIS this reduces to :
\begin{align}
    \nabla_x  \log \phi_{t}(x) &\approx \nabla_x\log \ptilde (x) + \frac{x}{T}
\end{align}
and for DDS
\begin{align}
    \nabla_x  \log \phi_{t}(x) &\approx \nabla_x\log \ptilde (\sqrt{1-\lambda_t} x) + \frac{\sqrt{1-\lambda_t}x}{v^2}
\end{align}

In both cases we can see the general form  $ \nabla_x  \log  \phi_{t}(x) =\nabla_x \log  \ptilde (a_t x) + c_t x$ for some time varying coefficients $a_t,b_t$, and thus it seems like a reasonable inductive bias to employ the Langevin preconditioning,
\begin{align}
  f_\theta(\cdot, t) = \text{NN}_{1,\theta}(\cdot, t) + \text{NN}_{2, \theta}(t) \circ ( \nabla\log \ptarget(a_t\cdot) + b_t \cdot),
\end{align}
or more simply $f_\theta(\cdot, t) = \text{NN}_{1,\theta}(\cdot, t) + \text{NN}_{2, \theta}(t) \circ  \nabla\log \ptarget(\cdot)$ as typically done in many SDE based neural samplers.

\section{Additional Experiment Details}

\subsection{Evaluation Metrics}
In this paper, we evaluate the samples quality by ELBO, EUBO and MMD. 
The ELBO (Evidence Lower Bound) is a lower bound of the (log) normalization factor, reflecting how well the model is concentrated within each mode;
on the other hand, EUBO \citep[Evidence Upper Bound, ][]{blessingbeyond}
provides an upper bound, representing if the model successfully covers all modes.
\par
The MMD (Maximum Mean Discrepancy) measures the distributional discrepancy between the generated samples and the target distribution.
We base our MMD implementation on the code by \citet{chen2024diffusivegibbssampling} at \url{https://github.com/Wenlin-Chen/DiGS/blob/master/mmd.py}, using 10 kernels and fixing the \texttt{sigma = 100}.
\par
For all experiments, we evaluate ELBO, EUBO and MMD with 10000 samples.

\subsection{Hyperparameters}

\begin{table}[H]
\centering
   \captionof{table}{Hyperparameters used for experiments.\vspace{-6pt}}\label{tab:hyperparam}
    \begin{tabular}{@{}lccccc@{}}
    \toprule
    method & objective & prior & lr & precond & network size \\ \midrule
    \multirow{6}{*}{DDS} & rKL & $\mathcal{N}(0, 30^2I)$ &  5e-4 & LG & [64, 64] \\
    & LV & $\mathcal{N}(0, 30^2I)$ &  5e-4  & LG & [64, 64] \\
    & TB & $\mathcal{N}(0, 30^2I)$ &   5e-4 & LG & [64, 64]\\
   & rKL & $\mathcal{N}(0, 30^2I)$ &  5e-4 & - / $\log p_\text{target}$& [256, 256, 256, 256, 256] \\
    & LV & $\mathcal{N}(0, 30^2I)$ &  5e-4  & - / $\log p_\text{target}$& [256, 256, 256, 256, 256] \\
    & TB & $\mathcal{N}(0, 30^2I)$ &   5e-4 & - / $\log p_\text{target}$  & [256, 256, 256, 256, 256]\\\midrule
    \multirow{4}{*}{CMCD} & rKL & $\mathcal{N}(0, 30^2I)$ & 5e-4 & LG & [64, 64] \\
    & LV & $\mathcal{N}(0, 30^2I)$ & 5e-3 & LG & [64, 64]\\ 
    & TB & $\mathcal{N}(0, 30^2I)$ & 5e-4 & LG & [64, 64] \\ 
    & rKL & $\mathcal{N}(0, 30^2I)$ & 5e-4 & - & [256, 256, 256, 256, 256]  \\
    \bottomrule
    \end{tabular}
\end{table}

We summarize the hyperparameters for DDS and CMCD in \Cref{tab:hyperparam}.
These hyperparameters are chosen according to \citet{blessingbeyond}.
For PINN-based experiments shown in \Cref{tab:pinn}, we follow the hyperparameter used in NETS \citep{albergo2024nets}, including network size, learning rate and its schedule. etc.

\section{Additional Experimental Results}\label{appendix:visualize}
\newpage
\begin{figure}[H]
    \centering
    \includegraphics[width=0.9\linewidth, trim={100, 0, 150, 0}, clip]{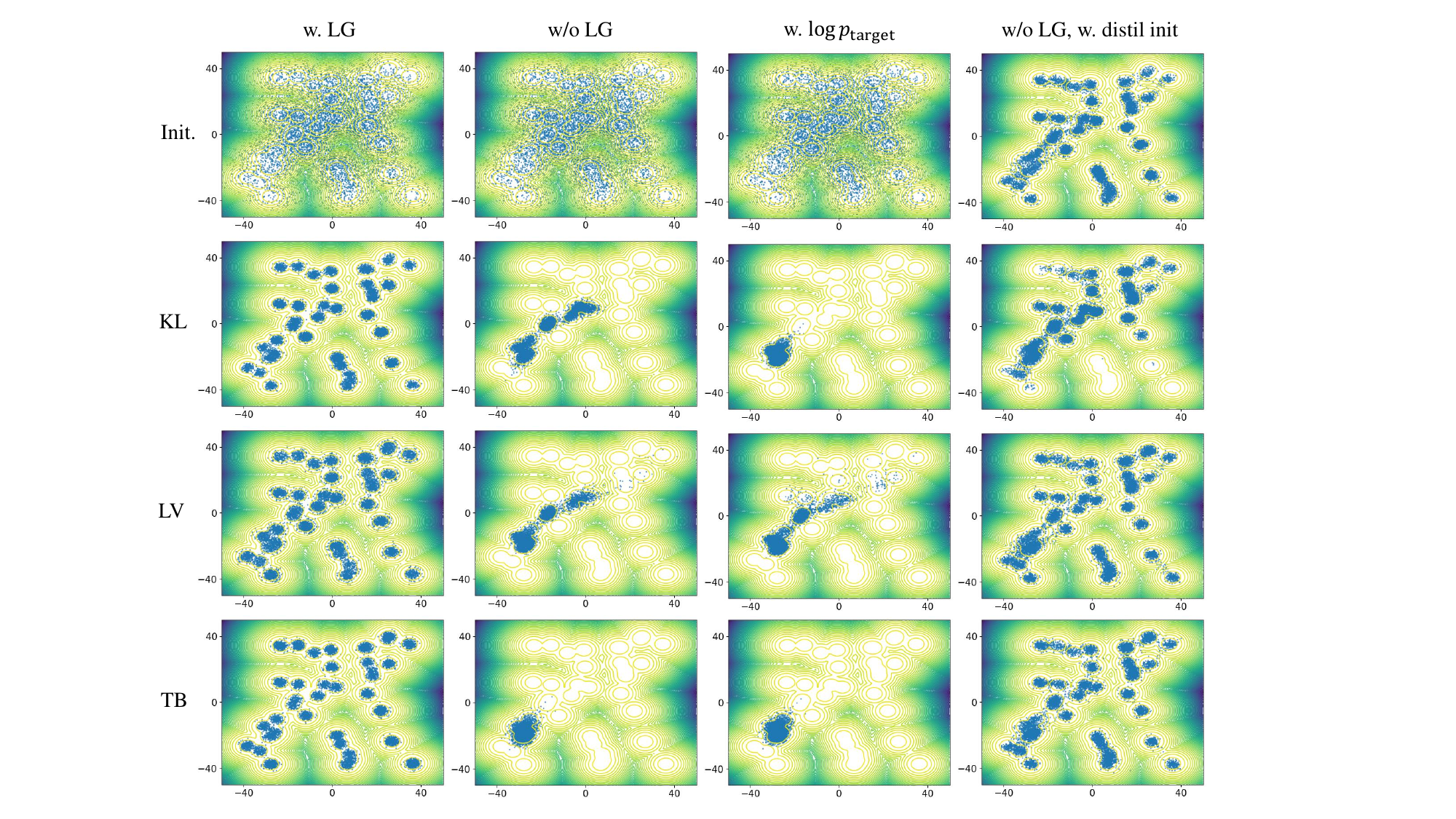}
    \caption{Sampled obtained by DDS with different settings.
    The first line shows the initialization. }
\end{figure}
\begin{figure}[H]
    \centering
    \includegraphics[width=0.9\linewidth, trim={100, 0, 150, 0}, clip]{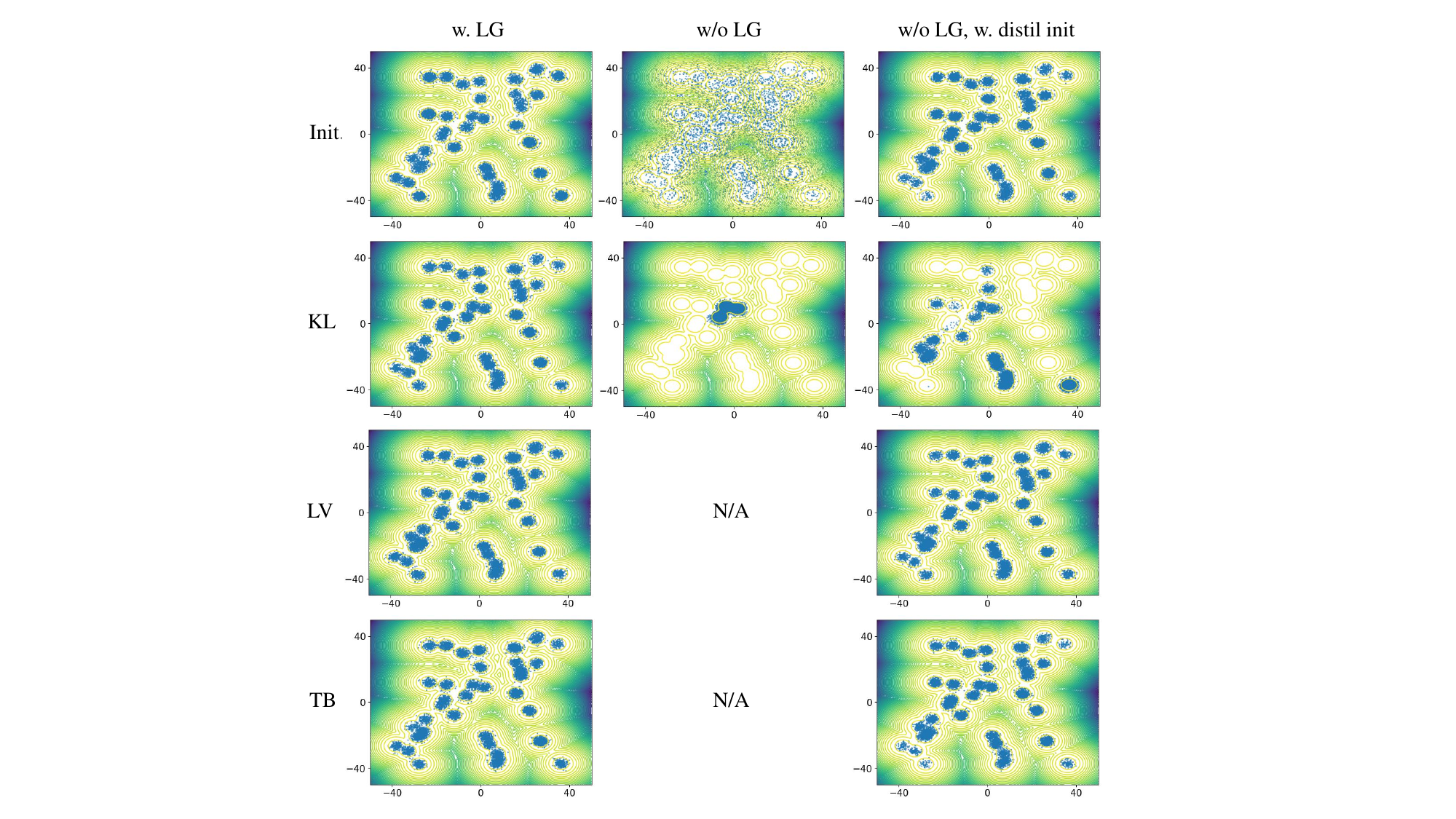}
    \caption{Sampled obtained by CMCD with different settings.
    The first line shows the initialization and N/A indicates diverging. 
    We can see when trained with Langevin preconditioning, we can see that CMCD already captures modes after initialization.}
\end{figure}

\end{document}